\documentclass[lettersize,journal]{IEEEtran}
\usepackage{amsmath,amsfonts}
\usepackage{algorithmic}
\usepackage{algorithm}
\usepackage{array}
\usepackage[caption=false,font=normalsize,labelfont=sf,textfont=sf]{subfig}
\usepackage{textcomp}
\usepackage{stfloats}
\usepackage{url}
\usepackage{verbatim}
\usepackage{graphicx}
\usepackage{cite}
\usepackage{hyperref}       
\usepackage{xspace}

\usepackage[normalem]{ulem} 
\usepackage{diagbox} 
\usepackage{multirow}
\usepackage[table]{xcolor}
\usepackage{fontawesome}

\newcommand{\posetrackmethodname}{\textcolor{black}{ViTPoseTrack}\xspace}

\begin{document}

\title{
APTv2: Benchmarking Animal Pose Estimation and Tracking with a Large-scale Dataset and Beyond
}

\author{Yuxiang Yang, Yingqi Deng, Yufei Xu, Jing Zhang~\IEEEmembership{Senior Member,~IEEE}
\thanks{The project was supported by the National Natural Science Foundation of China (62376080), the Zhejiang Provincial Natural Science Foundation Key Fund of China (LZ23F030003), and the Fundamental Research Funds for the Provincial Universities of Zhejiang (GK239909299001-003).
Corresponding author: Jing Zhang (jing.zhang1@sydney.edu.au)
}
\thanks{Y. Yang and Y. Deng are with the School of Electronics and Information, Hangzhou Dianzi University, Hangzhou 310018, China. Y. Xu and J. Zhang are with the School of Computer Science, The University of Sydney, NSW 2006, Australia.
}
}


\markboth{Journal of \LaTeX\ Class Files,~Vol.~14, No.~8, August~2021}%
{Shell \MakeLowercase{\textit{et al.}}: A Sample Article Using IEEEtran.cls for IEEE Journals}


\maketitle

\begin{abstract}
Animal Pose Estimation and Tracking (APT) is a critical task in detecting and monitoring the keypoints of animals across a series of video frames, which is essential for understanding animal behavior. Past works relating to animals have primarily focused on either animal tracking or single-frame animal pose estimation only, neglecting the integration of both aspects. The absence of comprehensive APT datasets inhibits the progression and evaluation of animal pose estimation and tracking methods based on videos, thereby constraining their real-world applications. To fill this gap, we introduce APTv2, the pioneering large-scale benchmark for animal pose estimation and tracking. APTv2 comprises 2,749 video clips filtered and collected from 30 distinct animal species. Each video clip includes 15 frames, culminating in a total of 41,235 frames. Following meticulous manual annotation and stringent verification, we provide high-quality keypoint and tracking annotations for a total of 84,611 animal instances, split into easy and hard subsets based on the number of instances that exists in the frame. With APTv2 as the foundation, we establish a simple baseline method named \posetrackmethodname and provide benchmarks for representative models across three tracks: (1) single-frame animal pose estimation track to evaluate both intra- and inter-domain transfer learning performance, (2) low-data transfer and generalization track to evaluate the inter-species domain generalization performance, and (3) animal pose tracking track. Our experimental results deliver key empirical insights, demonstrating that APTv2 serves as a valuable benchmark for animal pose estimation and tracking. It also presents new challenges and opportunities for future research. The code and dataset are released at \href{https://github.com/ViTAE-Transformer/APTv2}{https://github.com/ViTAE-Transformer/APTv2}.
\end{abstract}


\section{Introduction}
\label{sec:introduction}

\IEEEPARstart{P}{ose} estimation is a fundamental task in computer vision that involves identifying and localizing body keypoints in an image. This task plays a crucial role in various vision applications~\cite{arac2019deepbehavior,wang2021deep,wu2021spatiotemporal,zhang2020empowering}, such as behavior understanding and action recognition. While there has been significant progress in human pose estimation, driven by the availability of large-scale human pose datasets~\cite{lin2014microsoft,li2018crowdpose,mpii}, the focus on animal pose estimation remains limited, particularly in the context of video-based estimation. However, video-based animal pose estimation is of paramount importance in understanding animal behavior and promoting wildlife conservation.

Several endeavors have been made to establish animal pose estimation datasets and greatly foster research in this field. Unfortunately, they have primarily concentrated on pose estimation for specific animal categories on individual images. For instance, datasets such as those featuring horses~\cite{mathis2021pretraining}, zebras~\cite{graving2019deepposekit}, macaques~\cite{labuguen2020macaquepose}, flies~\cite{pereira2019fast}, and tigers~\cite{tiger} collect and annotate keypoints for particular animal species, contributing significantly to the progression of pose estimation research for these animals. Nonetheless, due to the extensive appearance variance, behavioral differences, and shifts in joint distribution across various animal species as a result of evolution, the models trained on these datasets demonstrate limited performance when applied to unseen animal species, thereby resulting in substandard generalization performance. In an effort to further encourage research on animal pose estimation, datasets that encompass multiple animal species with keypoint annotations have been proposed, such as Animal-Pose~\cite{Cao_2019_ICCV} and AP-10K~\cite{yu2021ap}. Despite their considerable scale and diversity in terms of animal species, these datasets do not incorporate essential temporal information that facilitates pose-tracking research. On the other hand, some efforts have been made to facilitate the development of animal instance tracking and contribute to the improvement of animal behavior understanding, like the Animal Track~\cite{zhang2022animaltrack} or Animal Kingdom~\cite{ng2022animal}. However, they do not have pose annotations based on the videos but only the bounding box annotations. Such an omission leaves the recognition of poses in consecutive frames unexplored, which is crucial for better animal action recognition and beyond.

To fill this gap, we present APTv2, the first large-scale dataset with high-quality animal pose annotations from successive frames, catering to both animal pose estimation and tracking. APTv2 comprises 2,749 video clips amassed and filtered from 30 diverse animal species. It contains a total of 41,235 frames, with 15 frames sampled from each video. These animals can be further categorized into 15 distinct taxonomic families, which helps in assessing the inter-species and inter-family generalization capacity of pose estimation models. To gather high-quality data for annotation, we manually select videos from the YouTube dataset following the practice in YouTube-VOS~\cite{xu2018youtube}. The selected videos will have high resolution and diverse backgrounds. After that, we extract frames at specific intervals to eliminate redundancy and increase the temporal motion amplitude. Subsequently, we employed 18 proficient annotators to label the keypoints for each animal in each frame in accordance with the MS COCO labeling protocols~\cite{lin2014microsoft}. These labels were then manually cross-verified for accuracy. Additionally, each animal's trajectory across the videos is indicated with bounding boxes and unique instance IDs.  Consequently, APTv2 can support research on both single-frame pose estimation and tracking of animal movements in successive frames. 

Utilizing APTv2, we establish three benchmark tracks for evaluating state-of-the-art (SOTA) pose estimation methodologies~\cite{sun2019deep,xiao2018simple,yuan2021hrformer,xu2022vitpose,xu2021vitae}: (1) Single-Frame Animal Pose Estimation (SF track), (2) Low-data Training and Generalization (LT track), and (3) Animal Pose Tracking (APT track). In the SF track, we extensively assess the performance of representative Convolutional Neural Networks (CNN) and Vision Transformer (ViT)-based methods across various settings, including both inter- and intra-domain transfer learning. Besides, we also investigate the influence of different pre-training datasets, including ImageNet~\cite{deng2009imagenet}, MS COCO human pose estimation dataset~\cite{lin2014microsoft}, and AP-10k~\cite{yu2021ap}, on the transfer learning performance, respectively. In the LT track, we evaluate the inter-family domain generalization capacity of the pose estimation model, which is trained on images of all species from some families and then tested on images of seen or unseen families. In the APT track, a number of object trackers, including a customized one based on the plain vision transformer~\cite{dosovitskiy2020image}, are utilized to track animal instances, while representative pose estimation models detect the keypoints of the tracked instances, and their performance is evaluated accordingly. What's more, a simple yet effective \posetrackmethodname baseline is provided, which utilizes a shared backbone for feature extraction and task-specific heads for pose estimation and tracking. Comprehensive experiment settings and results are provided in Section~\ref{sec:experiment}, where we highlight the immense potential of vision transformers for both animal pose estimation and tracking, the benefits of knowledge transfer between human and animal pose estimation, the advantages of incorporating diverse animal species into animal pose estimation, as well as the scalability of involving large scale models in animal pose tracking with the proposed baseline method.

The main contribution of this paper is three-fold
\begin{enumerate}
    \item We introduce APTv2, the pioneering large-scale benchmark for animal pose estimation and tracking. Its vast scale, wide range of animal species, and plentiful annotations of keypoints, bounding boxes, and instance IDs within consecutive frames make it a robust testing ground for future research.
    \item We establish three demanding tasks, specifically SF, LT, and APT, rooted in APTv2. We then comprehensively benchmark representative pose estimation models utilizing both CNNs and vision transformers, yielding valuable insights.
    \item A new baseline method called \posetrackmethodname with shared backbone and specific task heads for animal pose tracking tasks is established. With its simple structure, we have demonstrated the scalability and simplicity of large-scale models on the APT task, expanding the application boundaries of large models.
\end{enumerate}

\section{Related Work}
\label{sec:Related}
\subsection{Pose Estimation}

\subsubsection{Human pose estimation datasets} 
Human pose estimation has experienced rapid development both in terms of datasets~\cite{lin2014microsoft,li2018crowdpose,mpii} and methodologies~\cite{wang2022uformpose,xie2023rpm,tang2023ftcm,gai2023spatiotemporal}. Notably, MPII~\cite{mpii} and MS COCO~\cite{lin2014microsoft} are two widely recognized large-scale benchmarks for human pose estimation. To evaluate their performance in more challenging cases such as occlusions and crowd scenes, OCHuman~\cite{ochuman} and CrowdPose~\cite{li2018crowdpose}, have been established. Despite these substantial contributions, temporal information, which is critical for understanding human behavior and facilitating action imitation from humans to robots, has been largely overlooked. To address this deficiency, several video-based pose estimation datasets, such as VideoPose~\cite{videopose}, YouTube Pose~\cite{charles2016personalizing}, J-HMDB~\cite{jhuang2013towards}, and PoseTrack~\cite{andriluka2018posetrack}, have been introduced. These resources have considerably aided the research into human pose estimation and tracking~\cite{xiao2018simple}. Due to the distinct variance between humans and animals, models trained on these datasets are hard to generalize well on wild animals.

\subsubsection{Animal pose estimation datasets}
Given the vast diversity of wild animal species, animal pose estimation presents greater challenges than its human counterpart, and its progress significantly trails behind. Recently, there has been growing interest in animal pose estimation due to the increasing demand for understanding animal behavior and enhancing wildlife conservation. Early works introduced single-category animal pose estimation datasets, such as horse~\cite{mathis2021pretraining}, zebra~\cite{graving2019deepposekit}, macaque~\cite{labuguen2020macaquepose}, fly~\cite{pereira2019fast}, and tiger~\cite{tiger}. However, models trained on these datasets exhibit limited generalization abilities due to the substantial differences in appearance and movement patterns between different animal species. To remedy this, datasets covering multiple animal species have been developed, including Animal-Pose~\cite{Cao_2019_ICCV}, Animal Kingdom~\cite{ng2022animal}, and AP-10K~\cite{yu2021ap}, the latter of which contains 10,015 annotated images from 23 animal families and 54 species. Yet, these datasets lack temporal annotations, inhibiting the development of animal pose tracking methods. 

\subsubsection{Pose estimation methods}
Pose estimation has significantly evolved, transitioning from CNN~\cite{xiao2018simple,papaioannidis2022fast} to ViTs~\cite{yuan2021hrformer,xu2021vitae}. The typical process involves estimating a target's pose from given instances, no matter for humans or animals. To precisely locate keypoints, early CNN-based methods tend to rely on high-resolution features via techniques like skip feature concatenation~\cite{newell2016stacked} or highway structures~\cite{sun2019deep}. SimpleBaseline~\cite{xiao2018simple} directly recovers these high-resolution features using a decoder of several deconvolution layers. As ViTs have exhibited superior performance across various vision tasks~\cite{zhang2022segvit,li2022exploring}, some methods aim to use transformers as enhanced decoders following the CNN backbone~\cite{Li_2021_CVPR,li2021tokenpose,yang2021transpose}. For example, TransPose~\cite{yang2021transpose} and TokenPose~\cite{li2021tokenpose} adapt attention structures to model the relationships between different keypoints. The other kinds of methods rely on transformers only to deal with pose estimation methods. For example, HRFormer~\cite{yuan2021hrformer} uses transformers to extract high-resolution features directly. ViTPose~\cite{xu2022vitpose} further adopts plain vision transformers as backbones and demonstrates its scalability on human pose estimation. Despite their superior performance, they need a separate tracking model for pose tracking tasks. In contrast, we propose a simple baseline method named \posetrackmethodname that reuses the backbone model trained for pose estimation in tracking, enjoying efficient memory footprint, minimal model design, and good scalability of model size.



\subsection{Visual Object Tracking}
Object tracking~\cite{li2019siamrpn,hu2023stdformer,zhou2023memory,ni2023efficient} is a fundamental research task in computer vision. To advance the research in animal tracking, Animal Track~\cite{zhang2022animaltrack} and Animal Kingdom~\cite{ng2022animal} are proposed, which have diverse animal species. However, they do not have keypoint annotations for consecutive frames and are not thus suitable for the APT task. 
One common approach to object tracking follows the ``tracking by detection'' routine. Given the current frame and subsequent frames, an object detector is first used to identify candidate objects in the subsequent frames and the detected results are then associated with the target in the current frame. These methods have delivered superior results in both multiple object tracking (MOT)~\cite{wojke2017simple,zhang2022bytetrack} and single object tracking (SOT)~\cite{bertinetto2016fully,zhang2020ocean,wang2021transformer}. However, the generalization capabilities of these trackers are limited, \textit{i.e.}, they can only track objects that belong to categories that the detectors support. This limitation restricts their use in animal pose tracking, where many animal species may not be seen during detector training. Another approach to object tracking follows the ``tracking by matching'' pipeline. This approach employs a Siamese network to extract features from the tracked targets in the previous frame and the search regions in the subsequent frame and then matches these features to locate the targets in the search region. In this paper, we primarily adopt the single object tracking method for animal instance tracking, as it does not make assumptions about target categories. 

\subsection{Comparison to the Conference Version}
A preliminary version of this paper is presented in \cite{yangapt}. This paper extends the previous study with three major improvements:
\begin{enumerate}
    \item We specifically crawl, clean, and annotate hard cases (\textit{e.g.}, videos with more than two instances in each frame) to enrich the original dataset and approach real-world scenarios better, increasing the number of instances from 53,006 to 84,611. The extended dataset makes it more suitable to evaluate the performance of existing pose estimation and tracking methods thoroughly.
    \item We split our APTv2 dataset into easy and hard subsets according to the difficulty level and systematically investigate the influence of using them for training and evaluation, obtaining useful insights. Moreover, we investigate the scalability of model size, especially regarding ViTs, for the first time on the APT task.
    \item We delve further into the design and training aspects of the simple baseline method named \posetrackmethodname, with particular emphasis on the size of the model. Despite its simplistic design, this method achieves remarkably high performance, benefiting from both less memory footprint and good scalability in terms of model size.
    
\end{enumerate}

\section{Dataset}
\label{sec:data}

\begin{table*}[htbp]
  \centering
  \scriptsize
  \caption{Comparison of different animal pose datasets.}
    \setlength{\tabcolsep}{0.015\linewidth}{\begin{tabular}{c|cccccccc}
    \hline
          & {\#Species} & {\#Family} & {\#Frame} & {\#Keypoint} & {\#Sequence} & {\#Instance} &{\#Background Type} & {Difficulty Level}\\
    \hline
    Horses-10~\cite{mathis2021pretraining}  & 1 & N/A & 8,100 & 22 & N/A & 8,110 & N/A & N/A \\
    Animal-Pose Dataset~\cite{Cao_2019_ICCV}  & 5 & N/A & 4,666 & 20 & N/A & 6,117 & N/A & N/A \\
    Animal kingdom~\cite{ng2022animal}  & 850 & 6 & 33,099 & 23 & N/A & N/A & 9 & N/A \\
    AP-10k~\cite{yu2021ap}  & 54 & 23 & 10,015 & 17 & N/A & 13,028 & N/A  & N/A \\
    Animal track~\cite{zhang2022animaltrack}  & 10 & N/A & 24,700 & N/A & 58 & 429,000 & N/A & N/A \\
    APT-36K~\cite{yangapt}  & 30 & 15 & 36,000 & 17 & 2,400 & 53,006  & 10 & N/A\\
    APTv2  & 30 & 15 & 41,235 & 17 & 2,749 & 84,611  & 10 & Easy / Hard\\
    \hline
    \end{tabular}%
  \label{tab:dataset}}%
\end{table*}%

\subsection{Data Collection and Organization}
The aim of APTv2 is to establish a comprehensive, large-scale benchmark for animal pose estimation and tracking within real-world scenarios, an area largely untouched by prior work. In order to gather high-quality data that accurately represent natural living scenarios for animals, we turn to real-world video platforms, namely YouTube, for the careful selection and filtering of video clips. These clips, numbering in the hundreds, showcase 30 diverse animal species within a range of environments, from zoos and forests to deserts, which are typical habitats for the featured species. Additionally, to address potential imbalances due to the varied movement speeds of different animals and differing frame frequencies across the video clips, we meticulously set individual frame sampling rates for each video, leading to their tailored subsampling. This process ensures noticeable differences in movement and posture for each animal, thereby avoiding outlier cases where some animals may appear almost stationary over a given period. After the subsampling process, each video retains 15 frames. Importantly, our process ensures that challenging scenarios such as truncation and occlusion are retained in the dataset. This feature enables models to be evaluated under these challenging conditions, further enhancing the value and applicability of APTv2.

After the video collection and cleaning stages, we further categorize the videos of 30 animal species into 15 families, following the taxonomy~\cite{yu2021ap}. Typically, animals from the same taxonomic rank exhibit similarities in their behavior patterns, anatomical keypoint distribution, and appearance. For instance, the gait of dogs and wolves is quite similar, given they both belong to the Canidae family. In contrast, the walking pattern of a zebra, which belongs to a different family, \textit{i.e.}, Equidae, differs significantly from them. By adhering to the taxonomic rank, we create the potential to easily expand our dataset by collecting and annotating more videos from the same or different species or families. In addition, this structured organization of the animal pose dataset suggests a possible way to improve the generalization capabilities of animal pose estimation and tracking models for rare animal species. Specifically, this can be achieved by collecting and annotating videos from more commonly encountered animals of the same taxonomic rank.

\subsection{Data Annotation}
To ensure high-quality annotations for each image in the APTv2 dataset, 18 thoroughly trained annotators participated in the annotation process. We manually split the collected data into easy and hard subsets based on the number of instances in each video\footnote{While estimating the pose of a single instance in a frame may sometimes prove difficult (\textit{e.g.}, due to motion blur), detecting and tracking animal poses in multi-instance frames present even greater challenges due to occlusions, scale variation, and appearance ambiguity. Consequently, we adopt the number of instances per frame as a straightforward criterion for dividing the dataset into easy and hard subsets.}. The annotators are first instructed to make annotations based on the easy set, where there is only one instance in each video. To further enhance the annotation quality, we performed a stringent cross-check, which was repeated three times during the labeling process. After that, the annotators with better annotation skills participated in the annotation of the hard subsets with multiple instances per frame, followed by another stringent cross-check. The entire data collection, cleaning, annotation, and checking process consumed approximately 2,000 person-hours. Eventually, we labeled a total of 41,179 frames\footnote{Note that there are only a few frames (specifically 56) without labels due to the absence of objects.}, adhering to the COCO labeling format, with 24,700 easy images and 16,479 hard frames. We annotated at most 17 keypoints for each animal instance, including two eyes, one nose, one neck, one tail, two shoulders, two elbows, two hips, two knees, and four paws, following the protocol established by~\cite{yu2021ap}. 
Besides the keypoint annotations, we labeled the background type for each frame from 10 classes, \textit{i.e.}, grass, city, and forest. In addition, we assigned a unique tracking ID to each individual animal instance across the video clips. We split the dataset into three disjoint subsets for training, validation, and test, following a 7:1:2 ratio for each animal species. It is worth mentioning that we adopted a video-level partitioning approach to prevent potential information leakage, as frames within the same video clip bear a high degree of similarity.

\subsection{Statistics of the APTv2 Dataset}

\begin{figure}[htbp]
    \centering
    \includegraphics[width=0.9\linewidth]{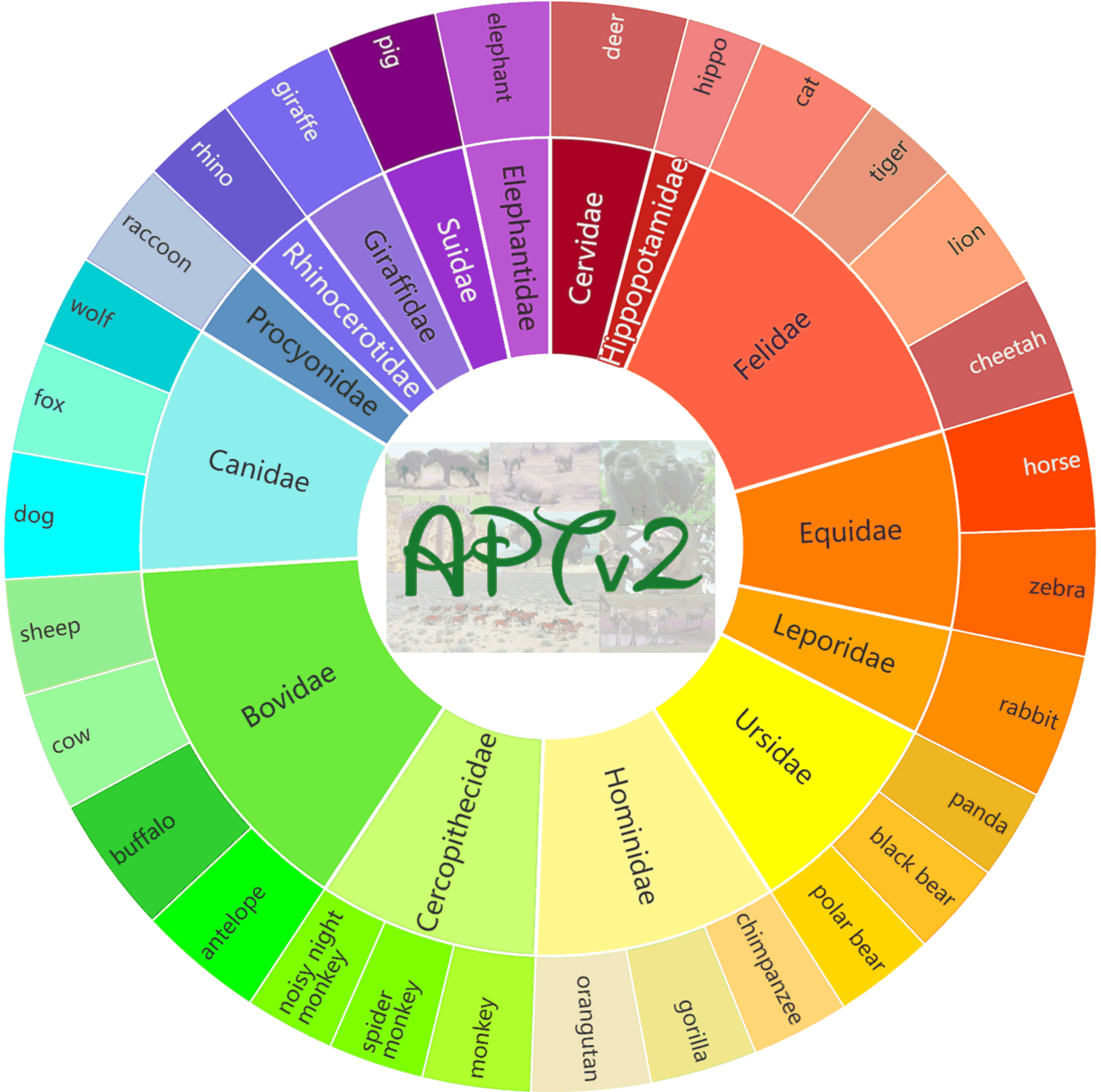}
    \caption{The taxonomic classification of animal families and their descendant species in our APTv2 dataset.}
    \label{fig:stastic}
\end{figure}

\begin{figure}[htbp]
    \centering
    \includegraphics[width=1\linewidth]{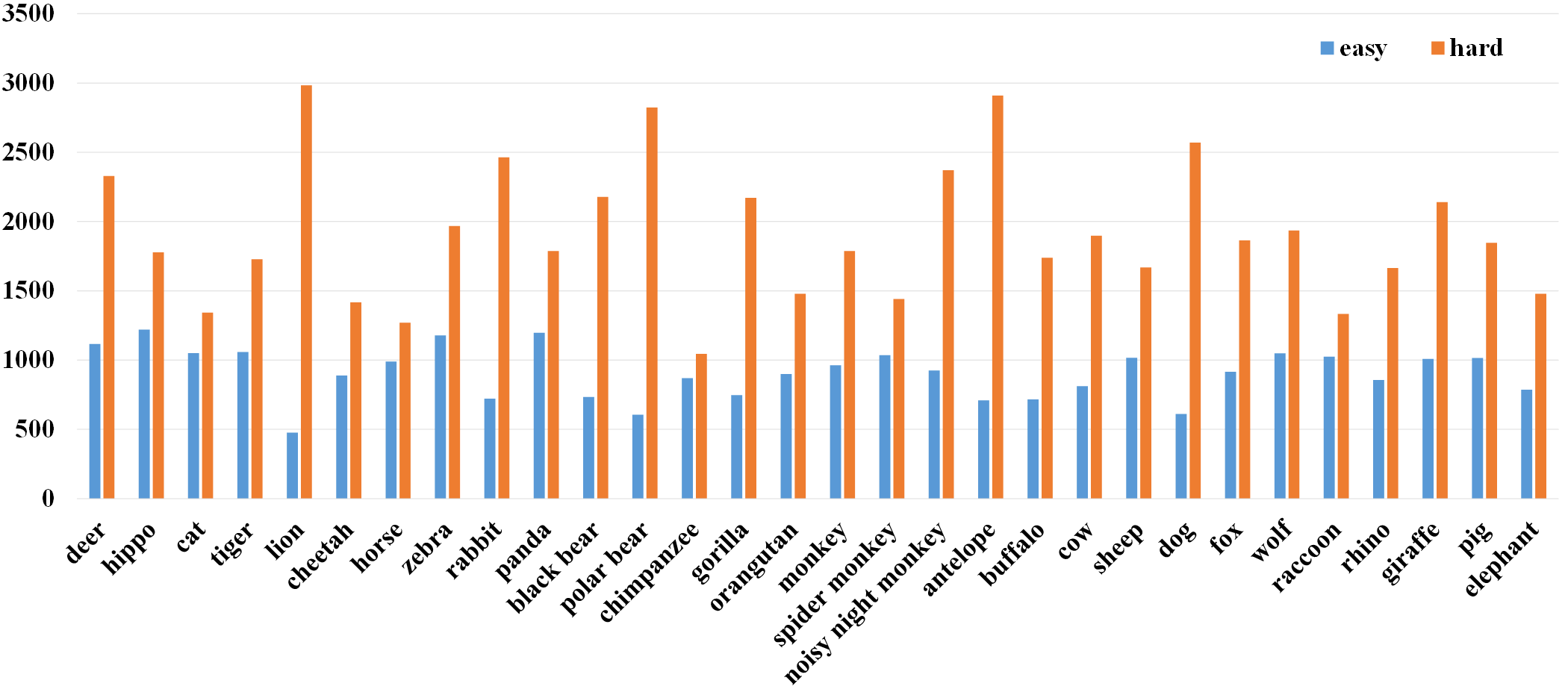}
    \caption{The number of instances per species in our APTv2 dataset.}
    \label{fig:stastic_instance}
\end{figure}

\begin{figure}[htbp]
    \centering
    \includegraphics[width=1\linewidth]{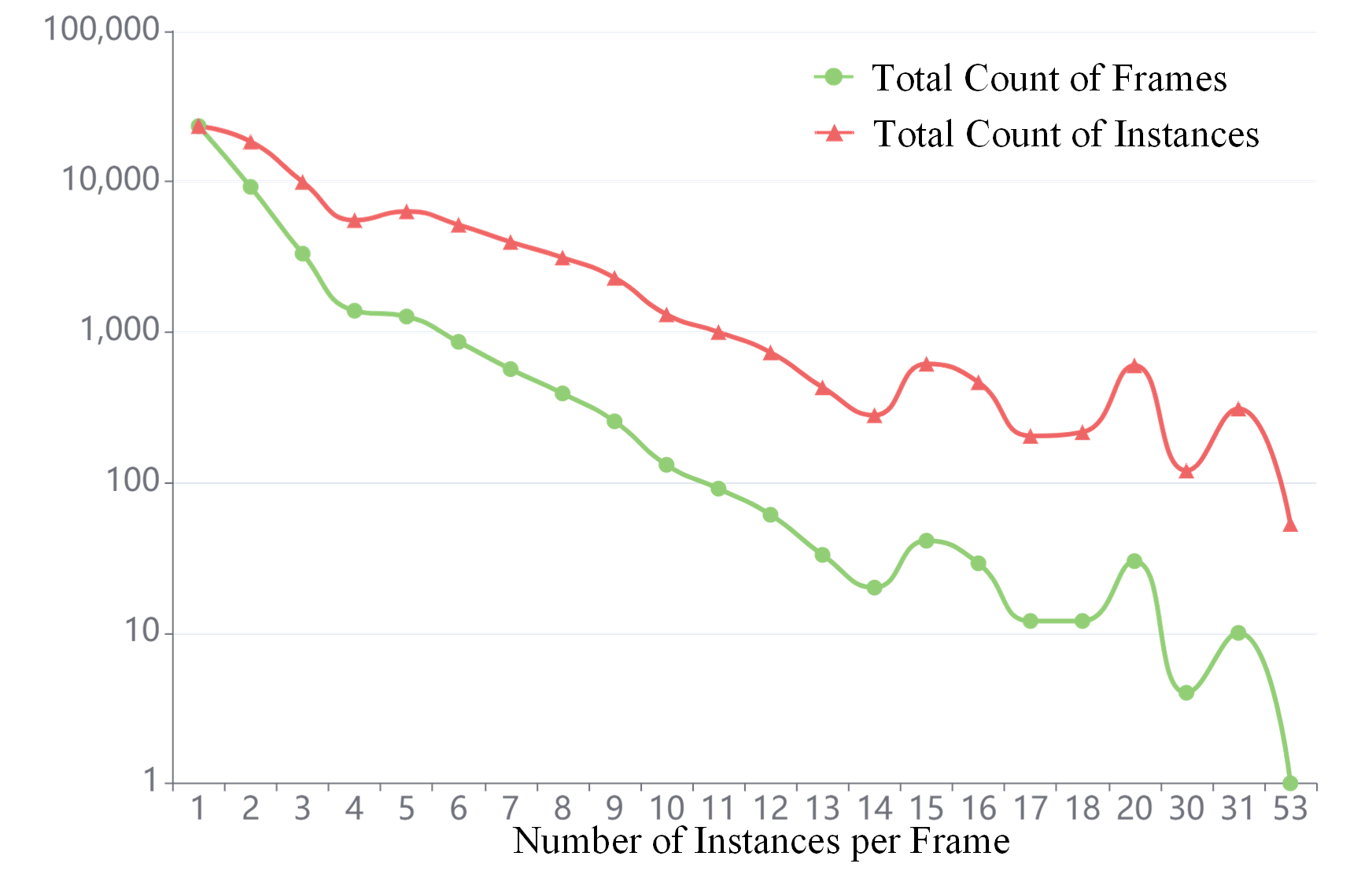}
    \caption{The total count of instances and frames with respect to the frequency of instances per frame in our APTv2 dataset.}
    \label{fig:stastic_frame}
\end{figure}

As demonstrated in Table~\ref{tab:dataset}, APTv2 covers 30 distinct animal species across 15 different families. It features a significant number of annotated frames with numerous annotated animal instances from a large collection of video clips. This comprehensive database surpasses prior animal pose estimation datasets, \textit{e.g.}, with much more instances than AP-10K dataset and much more species than Animal Pose and Animal Track dataset, thus establishing new challenges for tasks within the field. Besides, it is the first large-scale dataset that takes the animal pose track into consideration. Although the previous Animal Kingdom dataset covers both animal pose estimation and tracking, it treats the two tasks separately, which makes it not suitable for animal pose tracking tasks. To collect high-quality videos, the video clips used in APTv2 have been sourced from YouTube, spanning an array of topics such as documentaries, vlogs, and educational films, among others. Captured using different cameras and at varied shooting distances, these clips exhibit a diverse range of camera movement patterns. APTv2 incorporates images with ten different types of backgrounds, providing varied scenes for a holistic evaluation of animal pose estimation and tracking. 
Furthermore, we split our APTv2 dataset into easy and hard subsets according to the difficulty level, which is beneficial for systematically investigating the influence of using them for training and evaluation. As the first dataset suitable for both animal pose estimation and tracking, APTv2 fills an existing gap in this area and brings forth new challenges and opportunities for future research.



We also show the statistics of APTv2 to gain more insights about this dataset. As shown in Figure~\ref{fig:stastic_instance}, APTv2 contains more hard instances per animal species compared with the easy counterpart, posing more challenges on the animal pose estimation and tracking tasks. Such observation can be further validated by observing Figure~\ref{fig:stastic_frame}, where APTv2 exhibits a long-tail distribution concerning the frequency of instances per frame, \textit{i.e.}, certain frames contain a high number of animal instances, reaching up to 53. Moreover, these multi-instance frames make up a substantial portion of the dataset, containing a wealth of challenging instances.

\section{\posetrackmethodname Baseline}
We introduce a simple baseline method \posetrackmethodname in this section and demonstrate its good performance and scalability of model size on the APTv2 dataset.

\subsection{Architecture Design} 
\begin{figure*}
    \centering
    \includegraphics[width=0.9\linewidth]{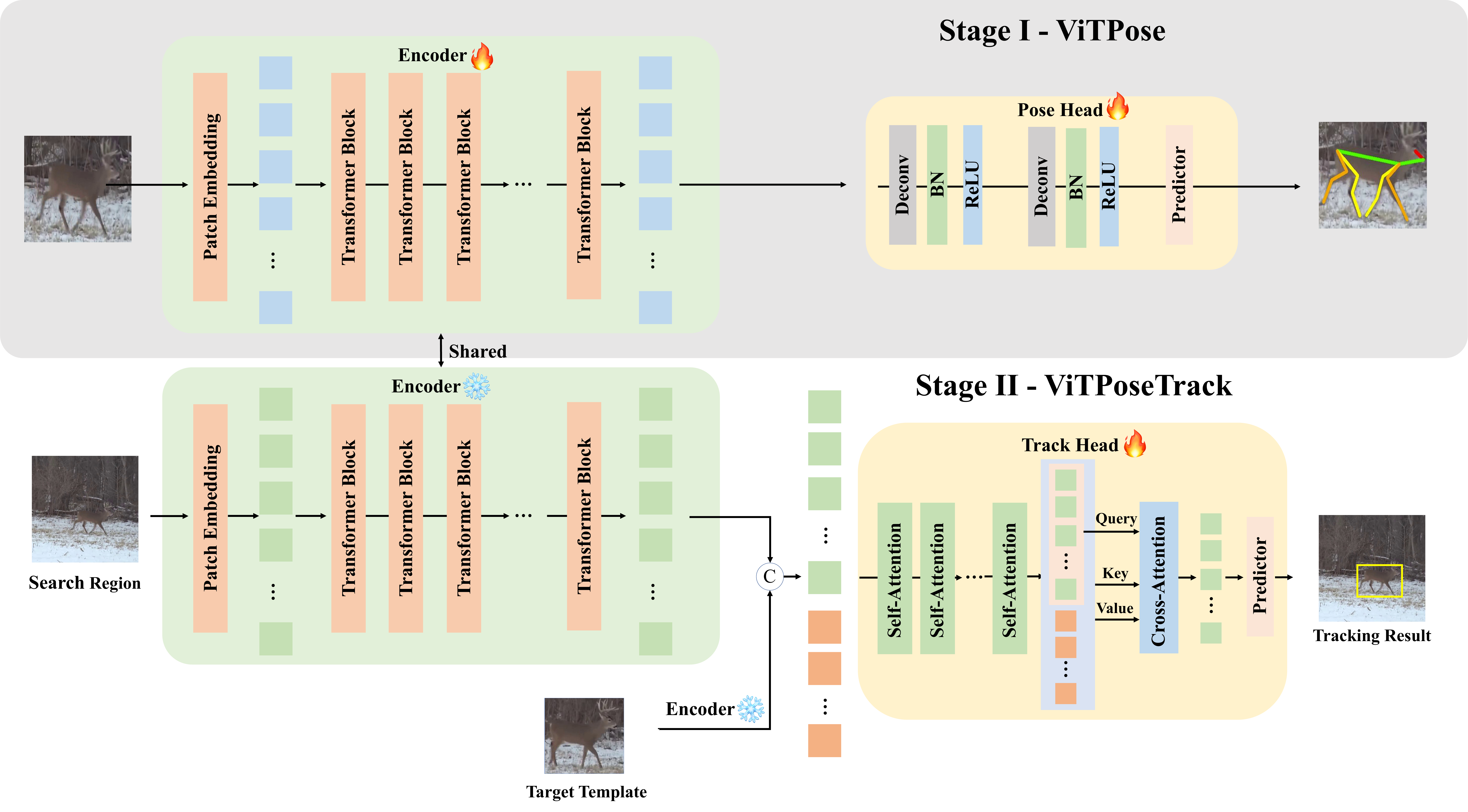}
    \caption{Architecture of the proposed \posetrackmethodname baseline model. The backbone encoder is shared for both pose estimation and tracking tasks and is only trained on the pose estimation task. The fire (snow) symbol denotes the corresponding module is trainable (frozen). }
    \label{fig:framework}
\end{figure*}

As shown in Figure~\ref{fig:framework}, \posetrackmethodname employs a simple encoder-decoder structure for both the animal pose estimation and tracking task. Specifically, a plain vision transformer is adopted as the encoder for feature extraction given the input image. The extracted feature is then fed into the task-specific head for pose estimation and tracking.

\subsubsection{Pose estimation head} The pose estimation head takes the classic structure as in SimpleBaseline~\cite{xiao2018simple}, which contains several deconvolution layers for feature upsampling and one projection layer to get the heatmap for each keypoint. After that, the locations of the corresponding keypoints are determined by finding the maximum values from each heatmap.

\subsubsection{Tracking head} In \posetrackmethodname, we follow the traditional tracking by matching pipeline. Specifically, the target object in the initial frame is utilized as the template for feature extraction. A larger search region, centered at the location of the tracked object in the preceding frame, is extracted from the current frame. The features of this search region are also extracted using the backbone network. Subsequently, these features of the template and the search region are concatenated and sent into an encoder for further feature processing. This encoder employs several transformer encoder layers. Based on the historical tracking trajectory, a set of additional motion tokens are generated. The processed features and these motion tokens are then input into a decoder to produce the prediction for the tracking results. This decoder is made up of several cross-attention layers, in which the tokens corresponding to the searched regions act as queries. The features from the template regions, search regions, and motion information serve as keys and values for the cross-attention layers. After being processed by the decoder, the search region tokens are used to predict the tracking results through a simple projection head.

\subsection{Training Schedule} 
In the training process, we first train the ViT encoder integrated with the pose estimation head end-to-end on the proposed APTv2 dataset. Following this, we freeze the ViT encoder and append the tracking head to it. In this setup, only the parameters of the tracking head are learnable during training on the tracking task. Both the box regression loss and the Intersection over Union (IoU) loss are utilized during the training phase of the tracking head, adhering to the design principles outlined in SwinTrack~\cite{lin2021swintrack}. This simplified design approach allows the backbone encoder to be shared across the two tasks, resulting in efficient memory footprint and inference speed. Experimental results validate that the features from the pose estimation task are distinctive enough to be successfully used in the tracking task.

\section{Experiment}
\label{sec:experiment}

\subsection{Representative Methods for Benchmarking}
We select several representative methods for the benchmark on the APTv2 dataset. We will briefly introduce them here.

\subsubsection{Pose estimation methods}
\textbf{SimpleBaseline}~\cite{xiao2018simple} is one of the representative CNN-based methods for pose estimation tasks. It follows the top-down pipeline and utilizes a ResNet~\cite{he2016deep} model as the backbone encoder for feature extraction. After that, three deconvolution layers are utilized for feature upsampling by 8 times and one projection layer is utilized to estimate the heatmap for each keypoint. \textbf{HRNet}~\cite{sun2019deep} argues that low-resolution features are harmful to obtain the precise location of the estimated keypoints. To overcome such limitations, it utilizes a highway structure to process high-resolution and low-resolution features simultaneously and gradually fuses them. The fused features are then directly fed into one projection layer to get the heatmap for each keypoint. \textbf{HRFormer}~\cite{yuan2021hrformer} is one representative transformer-based method, whose design is similar to that of HRNet~\cite{sun2019deep}. Specifically, it jointly processes high- and low-resolution features and utilizes window-based attention to reduce memory consumption. The high- and low-resolution features are gradually fused at each transformer stage. The high-resolution features are then processed by one projection layer to get the estimated heatmap. \textbf{ViTPose}~\cite{xu2022vitpose} focuses on estimating the appropriate keypoints with a plain vision transformer backbone. It utilizes the plain vision transformer for feature extraction and either a classic decoder or a simple decoder for upsampling feature maps and regressing the heatmaps. Due to its simple design, it is easy to benefit from the scalability of vision transformers, delivering superior results with efficient memory usage and fast execution.

\subsubsection{Tracking methods} 
\textbf{SiamRPN++}\cite{li2019siamrpn} utilizes ResNet\cite{he2016deep} as the backbone encoder for feature extraction. A Siamese structure is utilized with one branch for the template feature extraction and one branch to extract the feature of the search regions. Hierarchical features are utilized in the framework to gradually regress the location of the target object in incoming frames. \textbf{STARK}~\cite{yan2021learning} utilizes a transformer encoder-decoder structure for tracking. It first utilizes ResNet for feature extraction of the template and the search regions. The extracted features are then concatenated and fed into a transformer encoder for feature processing, where the processed features are then served as the key and values of the cross attention in the following transformer decoder. An extra track query is also fed into the transformer decoder to serve as the query and utilized to decode the location of the target in the incoming frames. \textbf{SwinTrack}~\cite{lin2021swintrack} gets rid of the ResNet for feature extraction and directly utilizes pure transformer structure for tracking. Specifically, it employs Swin Transformer~\cite{liu2021swin} for feature extraction of the template and search regions, respectively. The trajectory of the target object is also encoded into a motion token, which is then concatenated with the template and search region features to serve as the key and values of the cross-attention layer in the decoders. Different from STARK which needs extra track queries, SwinTrack directly regresses the location of the target based on the features from the search regions. \textbf{ViTTrack}~\cite{yangapt} refers to the tracking method that replaces the backbone of SwinTrack~\cite{lin2021swintrack}, originally based on Swin Transformer, with ViT-Base~\cite{dosovitskiy2020image}.

\subsection{Evaluation Metrics}
We adopt the average precision (AP) as the primary evaluation metric on the proposed APTv2 dataset. The average precision is defined according to the object keypoint similarity (OKS), for example, the loose metric AP$_{50}$ and the strict metric AP$_{75}$ are calculated by setting the OKS threshold to 0.5 and 0.75, respectively. AP is defined as the average value of the precision with a series of thresholds from 0.5 to 0.95~\cite{lin2014microsoft}.

\subsection{Implementation Details}\label{sec:implementaion}
To ensure a comprehensive evaluation for animal pose estimation and tracking, we benchmark various representative CNN-based and ViT-based pose estimation methods~\cite{sun2019deep,xiao2018simple,xu2022vitpose,yuan2021hrformer} on the newly proposed APTv2 dataset. Representative tracking methods~\cite{yan2021learning,li2019siamrpn,lin2021swintrack} and the proposed baseline method \posetrackmethodname are employed to obtain tracked boxes for the animal instances in the video clips. Based on APTv2, we establish three tracks: the SF track, LT track, and APT track, as outlined in Sec.~\ref{sec:introduction}. The pose estimation models are implemented using the MMPose~\cite{mmpose2020} codebase and are trained for 210 epochs, aligning with common practices in human/animal pose estimation tasks. The initial learning rate is set to 5e-4 and is decreased by a factor of 10 at the 170th and 200th epochs. For the training of the CNN-based methods, the Adam optimizer is utilized, and for the transformer-based methods, the AdamW optimizer is used. In the SF and LT tracks, we utilize ground truth bounding boxes, while in the APT track, we employ tracked boxes to evaluate the performance of the pose estimation methods, respectively. During the training of the baseline method \posetrackmethodname, we freeze the backbone encoder, and the tracking head is fine-tuned for 300 epochs using the AdamW optimizer, with a learning rate of 5e-4 and weight decay of 1e-4.

\subsection{Single-Frame Animal Pose Estimation (SF track)}
\subsubsection{Setting} In the SF track, we benchmark representative CNN-based and ViT-based pose estimation methods, such as SimpleBaseline~\cite{xiao2018simple}, HRNet~\cite{sun2019deep}, HRFormer~\cite{yuan2021hrformer}, and ViTPose~\cite{xu2022vitpose}. In addition to examining the impact of different backbone networks, we also delve into the benefits of various pre-training datasets, given the importance of pre-training in modern deep learning methods. We set up three settings to evaluate their performance, using network weights pre-trained on three different datasets: ImageNet-1K~\cite{deng2009imagenet}, MS COCO human pose estimation dataset~\cite{lin2014microsoft}, and the AP-10K dataset~\cite{yu2021ap}, respectively. 

For the ImageNet-1K pre-training, as is common practice, we initialize the backbones of pose estimation models with weights pre-trained on this dataset. We then fine-tune these models for additional 210 epochs on the APTv2 training set. It should be noted that we employ a fully supervised learning scheme on ImageNet-1K to acquire the pre-trained weights for the backbones used by SimpleBaseline, HRNet, and HRFormer. The ViT backbones in ViTPose are initialized with weights by self-supervised MAE pre-training~\cite{he2021masked}.

To evaluate the transferability of human pose estimation datasets to animal pose estimation tasks, we utilize the MS COCO dataset for pre-training. This decision is based on the idea that the keypoint definitions of quadrupeds are similar to those of humans. This approach allows us to leverage the existing large-scale datasets for human pose estimation to bridge the domain gap between the object-centric images in ImageNet-1K and animal images in APTv2. Accordingly, we pre-train both the CNN-based and ViT-based models on the MS COCO human pose estimation dataset for 210 epochs, and use the weights to initialize the backbone models for subsequent training on APTv2.

Lastly, to investigate the transferability between different animal pose estimation datasets, we adopt models pre-trained with abundant animal pose images and annotations from the AP-10K dataset~\cite{yu2021ap}. Specifically, we train these backbone models on the animal dataset for 210 epochs.

\begin{table*}[htbp]
  \centering
  \scriptsize
  \caption{Results (AP) of different models on the SF track of APTv2 with ImageNet-1K (IN1K)~\cite{deng2009imagenet}, MS COCO~\cite{lin2014microsoft}, and AP-10K~\cite{yu2021ap} pre-training, respectively. All, easy, and hard denote the entire validation set, and its easy and hard subsets, respectively.}
    \setlength{\tabcolsep}{0.015\linewidth}\begin{tabular}{c|c|cc|cc|cc|c}
    \hline
          & Pre-training & \multicolumn{2}{c|}{SimpleBaseline~\cite{xiao2018simple}} & \multicolumn{2}{c|}{HRNet~\cite{sun2019deep}} & \multicolumn{2}{c|}{HRFormer~\cite{yuan2021hrformer}} & ViTPose~\cite{xu2022vitpose} \\
          & dataset & ResNet-50 & ResNet-101 & HRNet-w32 & HRNet-w48 & HRFormer-S & HRFormer-B & VITPose-B \\
    \hline
    \multirow{3}[2]{*}{All} & IN1k~\cite{deng2009imagenet}  & 64.1  & 65.3  & 68.5  & 70.1  & 67.0  & 69.0  & 72.4 \\
            & AP-10K~\cite{yu2021ap} & 66.3  & 64.6  & 69.8  & 71.2  & 67.2  & 69.6  & 72.9 \\
            & COCO~\cite{lin2014microsoft}  & 67.8  & 68.1  & 70.1  & 71.7  & 69.5  & 69.7  & 72.5 \\
    \hline
    Easy  & COCO~\cite{lin2014microsoft}  & 75.9  & 75.5  & 78.9  & 80.1  & 77.6  & 77.8  & 79.7 \\
    \hline
    Hard  & COCO~\cite{lin2014microsoft}  & 65.4  & 65.9  & 67.7  & 69.4  & 67.3  & 67.4  & 70.4 \\
    \hline
    \end{tabular}%
  \label{tab:SF}%
\end{table*}%

\begin{table}[htbp]
  \centering
  \caption{Results (AP) on the APTv2 validation set with larger models. All, Easy, and Hard have the same meaning as in Table~\ref{tab:SF}.}
    \begin{tabular}{c|ccc}
    \hline
          & All   & Easy  & Hard \\
    \hline
    ViTPose-B~\cite{xu2022vitpose} & 72.5  & 79.7  & 70.4 \\
    ViTPose-L~\cite{xu2022vitpose} & 75.0  & 82.1  & 72.9 \\
    \hline
    \end{tabular}%
  \label{tab:SF_large}%
\end{table}%


\subsubsection{Results and analysis} The results are summarized in Table~\ref{tab:SF}, where we report the performances of representative methods on the easy subset, hard subset, and the entire validation set, respectively. A significant observation is the marked performance enhancement offered by the parallel-resolution design, which is evident in the comparison between the performances of HRNet and SimpleBaseline. Through the use of attention layers and suitable pre-training, transformer-based models can match performance levels akin to those without a parallel-resolution structure. For instance, the ViTPose base model achieves an Average Precision (AP) of 72.4, whereas the HRFormer base model, with ImageNet-1K pretraining, achieves an AP of 69.0.

The study further delves into the impact of various pre-training datasets. Notably, both CNN-based and ViT-based methods show performance improvements with human pose pre-training. The performance enhancements are demonstrated by an increase in AP from 64.1 to 67.8 for SimpleBaseline with a ResNet-50 backbone network, from 70.1 to 71.7 for HRNet-w48, and from 69.0 to 69.7 for HRFormer-B.

Interestingly, while the AP-10K pre-training results in smaller gains compared to MS COCO pre-training, it still significantly surpasses ImageNet pre-training. Despite the AP-10K dataset's diversity being substantially less than the MS COCO dataset, it suggests that leveraging animal pose datasets enhances the transfer performance of pose estimation methods on the proposed APTv2 dataset owing to the small domain gap. Furthermore, the scale of the pre-training dataset significantly impacts the performance of transfer learning. A larger pre-training dataset aids the baseline model in acquiring more discriminative features suitable for pose estimation tasks. Larger models can extract more valuable information from limited data, as further evidenced by the minimal performance gap between HRFormer base with AP-10K pre-training and MS COCO pre-training at 0.1 AP.

This observation is particularly noticeable in the comparison of MS COCO and AP-10K pre-training using the ViTPose base model, which achieved better results with AP-10K pre-training, \textit{e.g.}, 72.9 AP versus 72.5 AP. These results underscore the value of in-domain animal pose estimation datasets in pre-training and the extensive generalization capability of large-scale vision transformers.

Moreover, when compared to the easy validation set, the pose estimation models show lower performance on the hard subset. For example, AP results drop to 65.9 from 75.5 with ResNet-101, to 69.4 from 80.1 with HRNet-w48, to 67.4 from 77.8 with HRFormer base, and to 70.4 from 79.7 with ViTPose base. This underlines the need for improvements in current pose estimation methods, as most excel only in simple cases and struggle with complex scenarios involving occlusion, truncation, scale variation, appearance ambiguity, blur, etc. This insight emphasizes the challenge posed by the APTv2 dataset as a benchmark for current pose estimation methods.

To further examine the scalability of large vision transformers, we experimented with a larger ViTPose model with the ViT-Large backbone, pre-trained it on MS COCO, and fine-tuned it for 210 epochs on the APTv2 dataset. Despite this model having over 300M parameters, its training could be efficiently performed on low-end GPUs such as NVIDIA GTX 1080Ti and 3090, due to the efficient implementation of plain vision transformers. As seen in Table~\ref{tab:SF_large}, as the model size increases, the performance of ViTPose consistently improves. For instance, ViTPose-L achieves a gain of 2.5 AP over ViTPose-B, \textit{i.e.}, 75.0 AP versus 72.5 AP.

\subsection{Low-data Training and Generalization (LT track)}
\subsubsection{Setting} Collecting and annotating a substantial amount of images for each animal species, especially those that are endangered, represents a significant challenge. Consequently, evaluating the generalization capabilities of pose estimation methods becomes of paramount importance, particularly when the available training data is scarce. Considering the biological relationships among various animal species, we examine both the inter- and intra-species generalization abilities of the pose estimation models. Our evaluation encompasses two distinct settings. \textbf{1) Leave one out:} In this setting, we exclude all training samples that belong to a particular animal family from the training set. The models are then trained for 210 epochs using this refined training set and evaluated on the validation subsets split according to the selected families. Specifically, we have selected six animal families for this setting, \textit{i.e.}, Canidae, Felidae, Hominidae, Cercopithecidae, Ursidae, and Bovidae. We choose HRNet-w32 with MS COCO pre-training as the default model in this setting. \textbf{2) Low-data fine-tuning:} To account for the potential feasibility of annotating a limited number of instances for endangered animal species, we introduce a low-data fine-tuning setting to further assess the generalization capability of these pose estimation methods. From the proposed APTv2 dataset, we randomly select a specific number of images (\textit{e.g.}, 20, 40, 60, and 80) per animal species belonging to the above-selected families and use them collectively as the training set and adopt the same validation subsets as the above setting. The models are initialized using weights from models pre-trained on the AP-10K dataset. We choose HRNet-w32, ViTPose-B, and ViTPose-L in this setting.

\subsubsection{Results and analysis} 
\textbf{1) Leave one out:} The results on the ``leave one out'' setting are summarized in Table~\ref{tab:IS}. We find that the models demonstrate respectable generalization ability for the Canidae, Felidae, and Bovidae families when trained with instances from other animal families. For instance, a notable AP score of 66.8 is achieved for the Canidae family, as shown in the top-left cell of Table~\ref{tab:IS}, although the model has never seen the training data belonging to the Canidae family. This strong performance can be attributed to the shared features among Canidae, Ursidae, and Felidae families, as they all belong to the Carnivora order. This result underscores the potential advantage of training with similar species to mitigate the lack of specific animal species in the dataset.

However, for less common species that share fewer common features with those in the training set, the models exhibit weaker generalization ability. For example, the model trained without data from the Cercopithecidae family achieves an AP of just 45.0 for that same family. This underperformance can be mitigated by including relevant training data, leading to an average precision of 66.6 AP across all settings and a significant improvement exceeding 21 AP.

\begin{table}[htbp]
  \centering
  \scriptsize
  \caption{Results (AP) of HRNet-w32~\cite{sun2019deep} models on the ``Leave one out'' setting of APTv2. The performance on the unseen category is denoted with gray color and the others are seen categories. The Average (seen) score is calculated as the mean AP of seen categories. Fully supervised denotes that the model is trained with the whole training set as in the SF track. The performance gap between the mean AP on the seen categories and the AP of the unseen categories is denoted with $\Delta$.}
    \setlength{\tabcolsep}{0.006\linewidth}{\begin{tabular}{c|ccccccc}
    \hline
     \diagbox{training}{test} & {Canidae} & {Felidae} & {Hominidae} & {Cercopithecidae} & {Ursidae} & {Bovidae}  \\
    \hline
    \textit{w/o} Canidae  & \cellcolor[rgb]{ .847,  .847,  .847} 66.1  & 77.8 & 68.8 & 66.3 & 75.4 & 68.3  \\
    \textit{w/o} Felidae  & 74.4  & \cellcolor[rgb]{ .847,  .847,  .847}68.1 & 68.9  & 67.8  & 76.1  & 66.8 \\
    \textit{w/o} Hominidae  & 74.2  & 77.4  & \cellcolor[rgb]{ .847,  .847,  .847}49.7 & 65.5  & 75.9  & 67.4 \\
    \textit{w/o} Cercopithecidae  & 75.2  & 79    & 68.6  & \cellcolor[rgb]{ .847,  .847,  .847}45.0 & 76.2  & 67.7 \\
    \textit{w/o} Ursidae  & 74.2  & 77.9  & 69.9  & 66.4  & \cellcolor[rgb]{ .847,  .847,  .847}51.3 & 67.6 \\
    \textit{w/o} Bovidae  & 74.0  & 77.4  & 69.3  & 66.8  & 76.9  & \cellcolor[rgb]{ .847,  .847,  .847}59.5 \\
    \hline
    Average (seen) & 74.4 & 77.9 & 69.1 & 66.6 & 76.1 & 67.6 \\
    \hline
    $\Delta$ (seen - unseen) & 8.3 & 9.8 & 19.4 & 21.6 & 24.8 & 8.1 \\
    \hline
    \end{tabular}%
  \label{tab:IS}}%
\end{table}%

\begin{table}[htbp]
  \centering
  \scriptsize
  \caption{Results (AP) of HRNet-w32~\cite{sun2019deep} models on the ``Leave one out'' setting of APTv2, where only easy subset is used for training.}
    \setlength{\tabcolsep}{0.006\linewidth}{\begin{tabular}{c|ccccccc}
    \hline
     \diagbox{training}{test} & {Canidae} & {Felidae} & {Hominidae} & {Cercopithecidae} & {Ursidae} & {Bovidae}  \\
    \hline
    \textit{w/o} Canidae  & \cellcolor[rgb]{ .847,  .847,  .847} 55.9  & 68.0 & 63.2 & 60.8 & 61.8 & 55.9 \\
    \textit{w/o} Felidae  & 62.1  & \cellcolor[rgb]{ .847,  .847,  .847}59.5 & 64.6  & 60.9  & 61.7  & 56.0 \\
    \textit{w/o} Hominidae  & 62.6  & 70.1  & \cellcolor[rgb]{ .847,  .847,  .847}47.0 & 55.3  & 61.3  & 57.5 \\
    \textit{w/o} Cercopithecidae  & 62.6  & 69.1  & 62.9  & \cellcolor[rgb]{ .847,  .847,  .847}38.1 & 61.9  & 56.0 \\
    \textit{w/o} Ursidae  & 62.5  & 69.7  & 64.6  & 60.5  & \cellcolor[rgb]{ .847,  .847,  .847}45.2 & 55.7 \\
    \textit{w/o} Bovidae  & 61.8  & 69.5  & 65    & 59.9  & 63.4  & \cellcolor[rgb]{ .847,  .847,  .847}50.9 \\
    \hline
    Average (seen) & 62.3  & 69.3  & 64.1  & 59.5  & 62.0  & 56.2 \\
    \hline
    $\Delta$ (seen - unseen) & 6.4   & 9.8   & 17.1  & 21.4  & 16.8  & 5.3 \\
    \hline
    \end{tabular}%
  \label{tab:IS_easy}}%
\end{table}%


Further, we observe that images from the hard subset significantly contribute to enhancing the model's generalization performance, especially on the unseen families. As shown in Table~\ref{tab:IS} and Table~\ref{tab:IS_easy}, excluding hard images from the training set (\textit{i.e.}, training only with the easy subset) results in a significant performance drop, \textit{i.e.}, from 74.4 AP to 62.3 AP for the seen setting and from 66.1 AP to 55.9 AP for the unseen setting, on the Canidae family. 
However, the gap between the performance on seen and unseen settings narrows slightly when trained solely on the easy subset, suggesting that the hard subset contributes slightly more to improving performance for in-domain animal species than unseen ones. 


\textbf{2) Low-data fine-tuning setting} Table~\ref{tab:LDT} shows the results. The performance of all the models steadily improves as the volume of training data increases, underscoring the importance of collecting and labeling more data in enhancing the performance of animal pose estimation. Additionally, the scale of the models correlates positively with their generalization abilities. For example, when trained with only 20 images per species, HRNet-w32 achieves a modest AP of 59.1 while ViTPose-B achieves a superior AP of 63.5, surpassing the performance of HRNet-w32 even when the latter is trained with 80 images per species. Remarkably, the larger model ViTPose-L achieves an AP of 63.2 without using any training data from APTv2 for fine-tuning. Furthermore, with just 80 images per species for training (\textit{i.e.}, a total of 1,600 images), ViTPose-L obtains an AP of 71.6, outperforming the HRNet-w32 model trained with all the training data (\textit{i.e.}, 58,029 images) in the supervised setting, \textit{i.e.}, 71.3 AP. It validates the much better data efficiency (\textit{i.e.}, 35$\times$) of large models. These observations suggest a promising strategy for enhancing model generalization with limited data, \textit{i.e.}, scaling up the model size. While this finding aligns with results obtained from other domains like image classification, it represents a novel observation within the context of animal pose estimation.

\begin{table*}[htbp]
  \centering
  \scriptsize
  \caption{Results (AP) of HRNet-w32~\cite{sun2019deep}, ViTPose-B~\cite{xu2022vitpose}, ViTPose-L~\cite{xu2022vitpose} in the ``low-data fine-tuning'' setting.}
    \setlength{\tabcolsep}{0.011\linewidth}\begin{tabular}{c|c|cccccc|c}
    \hline
          & \# Training Samples per Species & Canidae & Felidae & Hominidae & Cercopithecidae & Ursidae & Bovidae & Average \\
    \hline
          & 0     & 59.6  & 64.5  & 42.2  & 38.6  & 51.6  & 58.7  & 52.5 \\
          & 20    & 63.3  & 68.4  & 53.5  & 50.2  & 58.9  & 60.2  & 59.1 \\
    HRNet-w32~\cite{sun2019deep} & 40    & 64.5  & 68.8  & 54.3  & 51.9  & 61.0    & 60.4  & 60.2 \\
          & 60    & 64.2  & 70.7  & 55.5  & 52.6  & 63.7  & 60.3  & 61.2 \\
          & 80    & 65.3  & 68.9  & 56.5  & 53.6  & 63.3  & 61.9  & 61.6 \\
          & All & 73.5 & 78.4 & 68.8 & 65.0 & 75.5 & 66.6 & 71.3 \\
    \hline
          & 0     & 63.9  & 65.5  & 47.5  & 51.0    & 59.0    & 59.0    & 57.7 \\
          & 20    & 67.1  & 69.3  & 59.9  & 59.1  & 63.6  & 62.2  & 63.5 \\
    ViTPose-B~\cite{xu2022vitpose} & 40    & 68.5  & 69.4  & 61.4  & 59.7  & 65.8  & 61.5  & 64.4 \\
          & 60    & 68.1  & 71.1  & 61.6  & 59.2  & 67.4  & 62.5  & 65.0 \\
          & 80    & 68.9  & 71.2  & 64.3  & 62.2  & 67.6  & 62.7  & 66.2 \\
    \hline
          & 0     & 66.0    & 69.1  & 59.9  & 61.0    & 62.4  & 61.0    & 63.2 \\
          & 20    & 70.0    & 73.6  & 69.3  & 68.3  & 67.6  & 64.6  & 68.9 \\
    ViTPose-L~\cite{xu2022vitpose} & 40    & 71.4  & 73.0    & 69.5  & 68.2  & 68.6  & 64.2  & 69.2 \\
          & 60    & 70.6  & 75.6  & 71.3  & 69.1  & 71.0    & 64.5  & 70.4 \\
          & 80    & 73.6  & 75.2  & 73.2  & 70.0    & 72.1  & 65.6  & 71.6 \\
    \hline
    \end{tabular}%
  \label{tab:LDT}%
\end{table*}%

\begin{table*}[htbp]
  \centering
  \scriptsize
  \caption{The tracking results of different models on the APTv2 validation set.}
    \setlength{\tabcolsep}{0.008\linewidth}{\begin{tabular}{c|cc|cc|cc|cc|c}
    \hline
          & \multicolumn{2}{c|}{SimpleBaseline~\cite{xiao2018simple}} & \multicolumn{2}{c|}{HRNet~\cite{sun2019deep}} & \multicolumn{2}{c|}{HRFormer~\cite{yuan2021hrformer}} & \multicolumn{2}{c|}{ViTPose~\cite{xu2022vitpose}} &  \\
          & ResNet-50 & ResNet-101 & HRNet-w32 & HRNet-w48 & HRFormer-S & HRFormer-B & VITPose-B & ViTPose-L & Average \\
    \hline
    SiamRPN++~\cite{li2019siamrpn} & 64.4  & 64.8  & 66.9  & 68.6  & 66.4  & 66.6  & 69.0  & 71.2  & 67.2 \\
    STARK~\cite{yan2021learning} & 63.5  & 63.7  & 65.9  & 67.2  & 65.3  & 65.6  & 67.5  & 69.5  & 66.0 \\
    SwinTrack~\cite{lin2021swintrack} & 65.0  & 65.2  & 67.7  & 68.8  & 66.7  & 66.9  & 69.3  & 71.4  & 67.6 \\
    ViTTrack~\cite{yangapt} & 65.9  & 66.0  & 68.2  & 69.5  & 67.6  & 67.7  & 70.2  & 72.2  & 68.4 \\
    \hline
    \posetrackmethodname-B & 65.3  & 65.6  & 67.9  & 69.1  & 66.9  & 67.1  & 69.7  & 71.9  & 67.9 \\
    \posetrackmethodname-L & 65.9  & 66.0  & 68.3  & 69.6  & 67.5  & 67.6  & 70.2  & 72.4  & 68.4 \\
    \hline
    Average & 65.0  & 65.2  & 67.5  & 68.8  & 66.7  & 66.9  & 69.3  & 71.4  & \\
    \hline
    \end{tabular}}%
  \label{tab:track}%
\end{table*}%

\begin{table*}[htbp]
  \centering
  \scriptsize
  \caption{The animal pose tracking results of different models on the easy validation subset.}
    \setlength{\tabcolsep}{0.012\linewidth}\begin{tabular}{c|cc|cc|cc|cc}
    \hline
          & \multicolumn{2}{c|}{SimpleBaseline~\cite{xiao2018simple}} & \multicolumn{2}{c|}{HRNet~\cite{sun2019deep}} & \multicolumn{2}{c|}{HRFormer~\cite{yuan2021hrformer}} & \multicolumn{2}{c}{ViTPose~\cite{xu2022vitpose}} \\
          & ResNet-50 & ResNet-101 & HRNet-w32 & HRNet-w48 & HRFormer-S & HRFormer-B & VITPose-B & ViTPose-L \\
    \hline
    SiamRPN++~\cite{li2019siamrpn} & 74.0    & 73.7  & 77.4  & 78.5  & 75.8  & 76.6  & 78.1  & 80.7 \\
    STARK~\cite{yan2021learning} & 75.2  & 75.0    & 78.5  & 79.6  & 76.8  & 77.4  & 79.1  & 81.4 \\
    SwinTrack~\cite{lin2021swintrack} & 75.5  & 75.3  & 78.8  & 79.8  & 76.9  & 77.5  & 79.6  & 81.9 \\
    ViTTrack~\cite{yangapt} & 75.7  & 75.3  & 78.6  & 79.8  & 77.0    & 77.7  & 79.4  & 81.6 \\
    \hline
    \posetrackmethodname-B & 75.7  & 75.5  & 78.6  & 79.8  & 77.0    & 77.6  & 79.7  & 82.0 \\
    \posetrackmethodname-L & 75.7  & 75.1  & 78.6  & 79.6  & 76.9  & 77.6  & 79.6  & 81.9 \\
    \hline
    \end{tabular}%
  \label{tab:track_easy}%
\end{table*}%

\begin{table*}[htbp]
  \centering
  \scriptsize
  \caption{The animal pose tracking results of different models on the hard validation subset.}
    \setlength{\tabcolsep}{0.012\linewidth}\begin{tabular}{c|cc|cc|cc|cc}
    \hline
          & \multicolumn{2}{c|}{SimpleBaseline~\cite{xiao2018simple}} & \multicolumn{2}{c|}{HRNet~\cite{sun2019deep}} & \multicolumn{2}{c|}{HRFormer~\cite{yuan2021hrformer}} & \multicolumn{2}{c}{ViTPose~\cite{xu2021vitae}} \\
          & ResNet-50 & ResNet-101 & HRNet-w32 & HRNet-w48 & HRFormer-S & HRFormer-B & VITPose-B & ViTPose-L \\
    \hline
    SiamRPN++~\cite{li2019siamrpn} & 61.6  & 62.2  & 63.9  & 65.7  & 63.8  & 63.7  & 66.2  & 68.3 \\
    STARK~\cite{yan2021learning} & 59.8  & 59.9  & 61.9  & 63.1  & 61.6  & 61.5  & 63.6  & 65.5 \\
    SwinTrack~\cite{lin2021swintrack} & 61.8  & 62.1  & 64.3  & 65.4  & 63.8  & 63.7  & 66.2  & 68.1 \\
    ViTTrack~\cite{yangapt} & 62.9  & 63.2  & 65.3  & 66.4  & 64.9  & 64.8  & 67.5  & 69.4 \\
    \hline
    \posetrackmethodname-B & 62.2  & 62.4  & 64.8  & 65.8  & 63.9  & 64.0    & 66.5  & 68.8 \\
    \posetrackmethodname-L & 63.0    & 63.2  & 65.3  & 66.6  & 64.9  & 64.8  & 67.3  & 69.7 \\
    \hline
    \end{tabular}%
  \label{tab:track_hard}%
\end{table*}%

\subsection{Animal Pose Tracking (APT track)}
\subsubsection{Setting} In this track, we employ representative object trackers with both CNN-based and ViT-based backbones to track each animal instance throughout the video clips, using each animal's ground truth bounding box in the first frame. Upon obtaining the tracked bounding boxes, the pose estimation models trained on the APTv2 training set are then applied for animal pose tracking. We also utilize the average precision (AP) metric for evaluation purposes.
For our baseline model \posetrackmethodname, we provide both the base and large models for comparison. Note that the tracking results of \posetrackmethodname can also be used by other pose estimation models to compare the tracking performance of \posetrackmethodname and other trackers.

\subsubsection{Results and analysis} The results on the entire validation set are presented in Table~\ref{tab:track}. The results on the easy and hard validation subsets are displayed separately in Table~\ref{tab:track_easy} and Table~\ref{tab:track_hard}, respectively.

Moreover, the proposed \posetrackmethodname baseline method exhibits the best performance among all trackers, even with the backbone frozen (see the last column in Table~\ref{tab:track}). This validates the effectiveness of pose estimation methods in extracting discriminative features at both instance and keypoint levels, thus highlighting the excellent generalization capability of large-scale ViT models. In addition, the ViTTrack model, with a trainable backbone, records a slightly better performance than ViTPoseTrack-B with the same backbone which is however frozen, as shown by comparing the 4$th$ and 5$th$ rows in Table~\ref{tab:track}.
Interestingly, the two configurations show comparable performance on the easy subset (Table~\ref{tab:track_easy}), while ViTTrack outperforms ViTPoseTrack-B by 0.8 average AP on the hard subset (Table~\ref{tab:track_hard}). This observation suggests that the APTv2 dataset provides a challenging benchmark for tracking methods, particularly within the hard subset. To further increase the model size of \posetrackmethodname, our \posetrackmethodname-L model with a ViT-Large backbone achieves the best performance, even if its backbone has not been trained on the tracking task.

\section{Conclusion}
\label{sec:con}
We introduce APTv2, covering 30 diverse animal species and 41,235 annotated frames from challenging real-world videos. This dataset serves as the first comprehensive resource for animal pose estimation and tracking. Leveraging APTv2, we evaluate state-of-the-art pose estimation methods in three distinct settings: single-frame animal pose estimation, low-data training and generalization, and animal pose tracking, with a novel baseline method \posetrackmethodname. Our extensive experiments highlight the advantages of inter- and intra-domain pre-training for accurate animal pose estimation. Furthermore, we emphasize the promising potential of involving large-scale pre-training models, e.g., plain vision transformers, in animal pose tracking tasks, as well as the importance of collecting and annotating diverse animal species' keypoints. By making APTv2 accessible, we aim to foster new avenues of research in animal pose estimation and tracking.

{
\bibliographystyle{ieee}
\bibliography{main}
}


 





\end{document}